\documentclass[conference]{IEEEtran}
\usepackage[T1]{fontenc}
\IEEEoverridecommandlockouts
\usepackage{cite}
\usepackage{amsmath,amssymb,amsfonts}
\usepackage{algorithmic}
\usepackage{graphicx}
\usepackage{hyperref}
\usepackage{textcomp}
\usepackage{xcolor}
\usepackage{caption}
\def\BibTeX{{\rm B\kern-.05em{\sc i\kern-.025em b}\kern-.08em
    T\kern-.1667em\lower.7ex\hbox{E}\kern-.125emX}}
\begin{document}

\title{Synthesizing Images on Perceptual Boundaries of ANNs for Uncovering Human Perceptual Variability on Facial Expressions}

\author{
\IEEEauthorblockN{ Haotian Deng\textsuperscript{1, *}, Chi Zhang\textsuperscript{1, *}, Chen Wei\textsuperscript{1, 2, \dag}, Quanying Liu\textsuperscript{1, \dag}}
\IEEEauthorblockA{\textsuperscript{1}\textit{Department of Biomedical Engineering, Southern University of Science and Technology, Shenzhen, China}\\ \textsuperscript{2}\textit{University of Birmingham, Birmingham, United Kingdom} \\
\{12313204, 12210315, 12150103\}@mail.sustech.edu.cn; liuqy@sustech.edu.cn}
\thanks{\textsuperscript{*} These authors contributed equally to this work.}
}

\renewcommand{\thefootnote}{} 

\twocolumn[{%
\renewcommand\twocolumn[1][]{#1}%

\maketitle
\begin{center}
    \includegraphics[width=\textwidth]{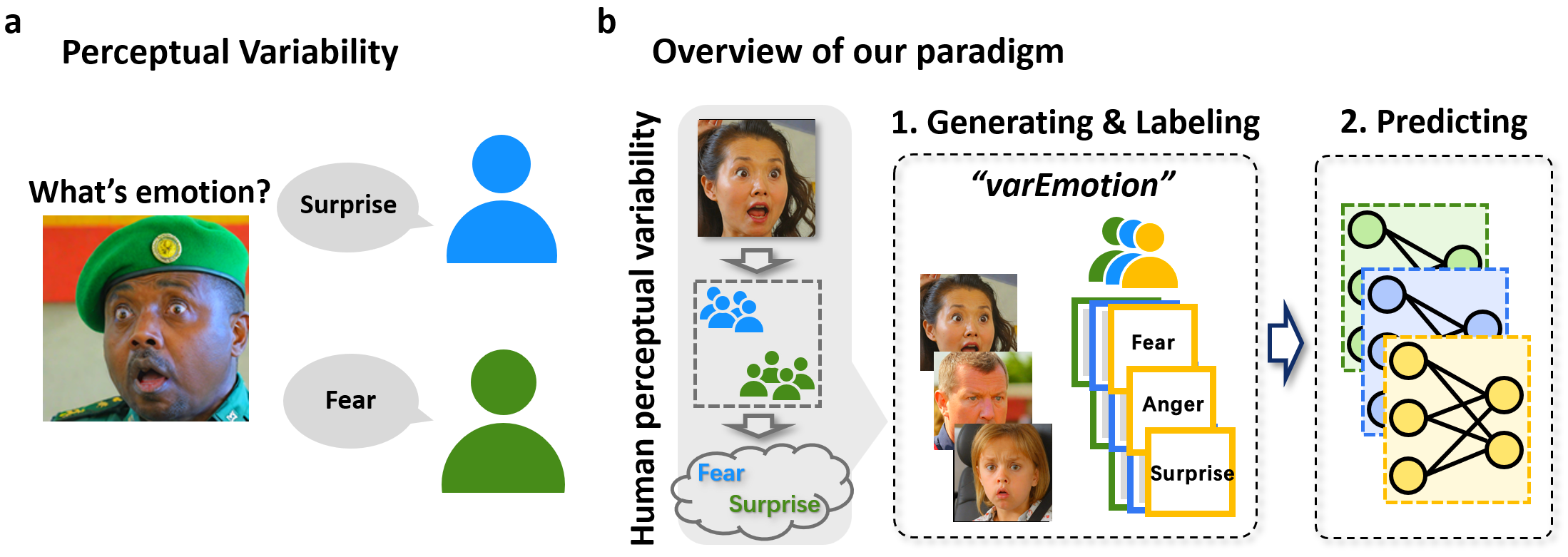}
    \captionof{figure}{\textbf{Overview of our paradigm.} (a) \textbf{Motivation:} An example of perceptual variability. 
    (b) Our approach consists of two main components: \textbf{1. Generating \& labeling}: Sampling images from ANN decision boundaries and using them in human behavioral experiments to construct the high-variability dataset varEmotion; \textbf{2. Predicting}: Finetuning models with human behavioral data to align them with human perceptual variability at the group and individual levels, enhancing behavior prediction accuracy.}
    \label{fig:overview}
\end{center}
}]

\footnotetext{\textsuperscript{*} Equal contribution.}
\footnotetext{\textsuperscript{\dag} Corresponding author.}

\begin{abstract}

A fundamental challenge in affective cognitive science is to develop models that accurately capture the relationship between external emotional stimuli and human internal experiences. While ANNs have demonstrated remarkable accuracy in facial expression recognition, their ability to model inter-individual differences in human perception remains underexplored. This study investigates the phenomenon of high perceptual variability—where individuals exhibit significant differences in emotion categorization even when viewing the same stimulus. Inspired by the similarity between ANNs and human perception, we hypothesize that facial expression samples that are ambiguous for ANN classifiers also elicit divergent perceptual judgments among human observers. To examine this hypothesis, we introduce a novel perceptual boundary sampling method to generate facial expression stimuli that lie along ANN decision boundaries. These ambiguous samples form the basis of the \textit{varEmotion} dataset, constructed through large-scale human behavioral experiments. Our analysis reveals that these ANN-confusing stimuli also provoke heightened perceptual uncertainty in human participants, highlighting shared computational principles in emotion perception. Finally, by fine-tuning ANN representations using behavioral data, we achieve alignment between ANN predictions and both group-level and individual-level human perceptual patterns. Our findings establish a systematic link between ANN decision boundaries and human perceptual variability, offering new insights into personalized modeling of emotional interpretation.
\end{abstract}

\begin{IEEEkeywords}
Perceptual Variability, Facial Expression Recognition, Emotion Perception, Human-AI Alignment
\end{IEEEkeywords}

\section{Introduction}


A core goal of affective cognitive science is to develop models that accurately capture the relationship between external emotional stimuli and human internal experiences. The advancement of artificial neural networks (ANNs) has significantly contributed to this goal, particularly as their latent representations have been shown to strongly correlate with human psychological representations \cite{wei2024cocog, wei2024cocog2, muttenthaler2022human, mahner2024dimensions, zheng2019revealing, hebart2020revealing, muttenthaler2022vice}.
This study focuses on a critical phenomenon: even when exposed to the same emotional stimuli, individuals may exhibit significant differences in their internal perceptual experiences. While such perceptual variability has been widely studied in complex cognitive tasks (e.g., aesthetic or moral judgments), individual differences in simpler visual decision tasks, such as facial expression recognition, have often been overlooked. As illustrated in Figure \ref{fig:overview}(a), when different individuals observe the same stimulus, they may categorize it as different emotions (e.g., “anger” vs. “fear”). However, this \textit{high perceptual variability} remains inadequately explored in this field, despite modern neural networks achieving remarkable accuracy in facial expression recognition \cite{goodwin2020face}. 
Inspired by the similarity between ANNs and human perception, we hypothesize that facial expression samples that are ambiguous for ANN classifiers are also difficult for human participants to recognize. These stimuli serve as key examples that elicit divergent perceptual judgments across individuals, highlighting systematic differences in emotional interpretation.

\begin{figure*}[htbp] 
    \centering
    \includegraphics[width=1\textwidth]{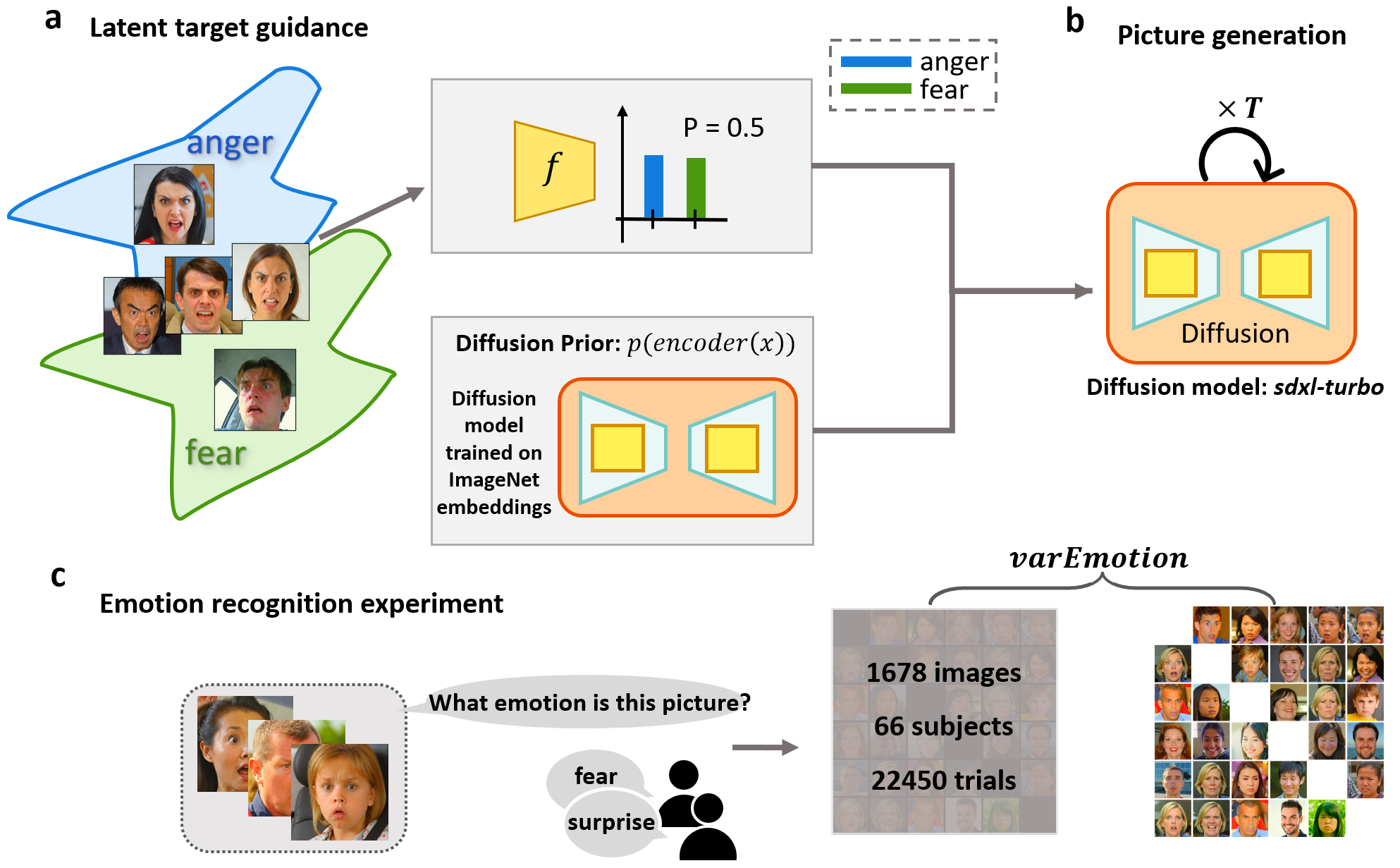} 
    \caption{\textbf{Generating images to elicit human perceptual variability.} (a) This example demonstrates how to generate embeddings by sampling from the perceptual boundaries of the expressions `anger' and `fear' in an ANN using the uncertainty guidance method. The goal of uncertainty guidance is to focus the ANN's prediction of the embeddings on `anger' and `fear'. The diffusion model follows the first stage generation process from CoCoG \cite{wei2024cocog}, taking the original image embeddings as a prior and guiding the denoising process toward the desired target direction. The ANN model used for guidance is an MLP pre-trained on image embeddings from the RAF-DB dataset, with an output of six channels representing the emotions [`surprise', `fear', `disgust', `happiness', `sadness', `anger'].
    (b) Two-stage image generation. We reference the two-stage generation method from DALL-E 2 \cite{ramesh2022hierarchical}, using the sdxl-turbo model without a classifier to generate images from the embeddings generated in the first stage. Only prompt guidance is applied during the second-stage generation.
    (c) Using the methods described above, we generated a series of images capable of inducing uncertainty in the ANN and used these images in a human experiment. In the experiment, human participants are shown the generated images and asked to choose the emotion of the face depicted in each image. A total of 1678 images were used, with 22450 trials conducted across 66 participants, resulting in the high perceptual variability dataset varEmotion.
    }
    \label{fig2}
\end{figure*}
Emerging methodologies using ANNs as perceptual probes offer promising research avenues. The discovery that imperceptible image perturbations alter both machine and human judgments \cite{veerabadran2023subtle} suggests shared computational principles in visual processing. Recent work by \cite{gaziv2024strong} further establishes that minimal stimulus modifications can induce perceptual conflicts across biological and artificial systems. Building on the conceptual framework of model metamers \cite{feather2023model}—stimuli equivalent for ANNs but distinguishable by humans—we develop a novel paradigm for facial expression analysis. \cite{golan2020controversial,golan2023testing} introduced \textit{controversial stimuli}, designed to elicit divergent judgments across models, further highlighting their misalignment with human perception. Our method directly links ANN decision boundaries to human perceptual variability through facial expression stimulus generation.

Our methodological framework is composed of three components:
\textbf{1. Generation of high emotional variability stimuli} 
Making use of the perceptual boundary sampling approach (as described in Sec.~\ref{sec:method}), we create a set of facial expressions along the decision boundaries of ANNs for six fundamental emotion categories. These stimuli retain their photorealistic authenticity due to the generative uncertainty constraints.
\textbf{2. Behavioral Validation}: We select images from the perceptual boundaries of ANNs and build the \emph{varEmotion} dataset via human behavioral experiments. This enables us to systematically record the inter-individual differences in emotional perception.
\textbf{3. Individual Alignment}: We achieve alignment of the models for perceptual variability at both the group and individual levels by fine - tuning ANN models with the utilization of human behavioral data.

Our key contributions are as follows:

(1) We engineered a sophisticated algorithm tailored to sample precisely on the classification boundaries of Artificial Neural Networks (ANNs) employed in facial expression recognition. By leveraging the unique characteristics of these boundaries, this algorithm generates samples that present formidable challenges to ANNs, causing them to struggle in reaching definitive decisions. These samples are of great value as they push the ANNs to their decision - making limits, enabling a deeper exploration of the network's performance and robustness in facial expression recognition scenarios.

(2) Through extensive large - scale behavioral experiments, we sampled data from the classification boundaries of ANNs and utilized the results to construct the varEmotion dataset. A comprehensive quantitative analysis was then carried out on this dataset. The findings clearly indicate that the samples which confound ANNs also substantially heighten the decision - making uncertainty among human subjects. This connection between ANN and human decision - making difficulties provides new insights into the shared cognitive processes underlying facial expression perception.

(3) We achieved a successful alignment of ANNs with human subjects at both the group and individual levels. Our in - depth analysis reveals that individuals display distinct and significant preferences when performing facial expression recognition tasks. Remarkably, these individual - specific preferences can be effectively learned and modeled using a relatively limited number of experimental samples. This discovery paves the way for the development of more personalized and accurate facial expression recognition systems.

\section{Related Works}


Researchers have widely employed ANN - generated synthetic images to explore human perceptual space. They've found disparities between model and human perception and refined generation techniques for greater influence on human cognition. \cite{golan2020controversial,golan2023testing} used controversial stimuli to show classification differences in neural networks. \cite{veerabadran2023subtle} showed that adversarial perturbations can affect both ANN classifications and human perceptual choices, indicating shared sensitivities. But \cite{gaziv2024strong} noted that standard ANN perturbations don't impact human perception, while robustified ANN models can generate low - norm perturbations that disrupt human percepts.
Some studies took different approaches. Feather, Nanda et.\cite{feather2023model,feather2019metamers,nanda2022measuring,nanda2023invariances} studied \textit{model metamers}, revealing mismatches between model activations and human recognition. \cite{fu2023dreamsim} introduced DreamSim, a metric using synthetic and human experimental data to better reflect human similarity judgments and fix flaws in traditional metrics. Recent work, like \cite{muttenthaler2024aligning,sundaram2024does}, aimed to align vision models with human perceptual representations by adding human - like concepts, improving alignment and performance.
For studying human perceptual variability, generated images must strongly influence human cognition. Since samples from ANN perceptual boundaries are often noisy, better methods are needed for natural - looking images. Machine - learning studies on adversarial examples and counterfactual explanations, such as \cite{jeanneret2023adversarial}, \cite{wei2024cocog2}, \cite{chen2023advdiffuser}, \cite{jeanneret2022diffusion}, \cite{vaeth2023diffusion}, and \cite{atakan2023dreamr}, use diffusion models with training - free guidance \cite{yu2023freedom,ma2023elucidating,yang2024guidance} as regularizers. This helps introduce prior distributions, enhancing image naturalness and their impact on human perception.

In the field of psychology, researchers have been searching for the most basic facial expressions. Many studies (\cite{ekman1978facial}, \cite{cordaro2018universals}, \cite{keltner2019emotional}, \cite{matsumoto2008facial}, \cite{jack2014dynamic}) suggest that there are six universal basic expressions, namely \textit{`surprise', `fear', `disgust', `happiness', `sadness', `anger'}. However, Snoek \cite{snoek2023testing}, Cordaro \cite{cordaro2018universals}, and others have pointed out that individuals differ in their perception and recognition of these expressions, and that these differences are related to cultural factors. Therefore, we propose that different individuals have different perceptual boundaries when recognizing facial expressions, and that ANNs also have specific perceptual boundaries for facial expression recognition. In the next phase of the study, we will sample the perceptual boundaries of expressions in ANNs and generate a series of facial images with uncertain expressions using a diffusion model. These images will be used in human experiments, where participants will identify the emotions corresponding to these images. Based on the feedback data from the participants, we will fine-tune the ANN to enable it to more accurately fit the human perceptual boundaries.

\section{Method}
\label{sec:method}
In this section, we introduce a method for generating pictures of human facial expression with high perceptual variability. We also conduct human experiment online and constructed the \textit{varEmotion} dataset: a facial dataset with high perceptual variability. Our goal is to generate images that evoke significant human perceptual variability and collect this variability by recording human perceptual judgments on the generated images.

\subsection{Generating Images on ANN perceptual boundary}
Many existing studies indicate that images which significantly affect the perception of artificial neural networks (ANNs) can also influence human perception (\cite{gaziv2024strong}, \cite{muttenthaler2022human}, \cite{veerabadran2023subtle}, \cite{wei2024cocog}). This suggests that ANNs may share a similar perceptual space with humans. Based on this, we propose that stimuli generated by sampling the perceptual boundaries of ANNs could similarly induce perceptual variability in humans, leading to differences in how individuals perceive and judge the stimuli.

\subsection{Facial expression recognition experiment}
\begin{figure}[t]
    \centering
    \includegraphics[width=1\linewidth]{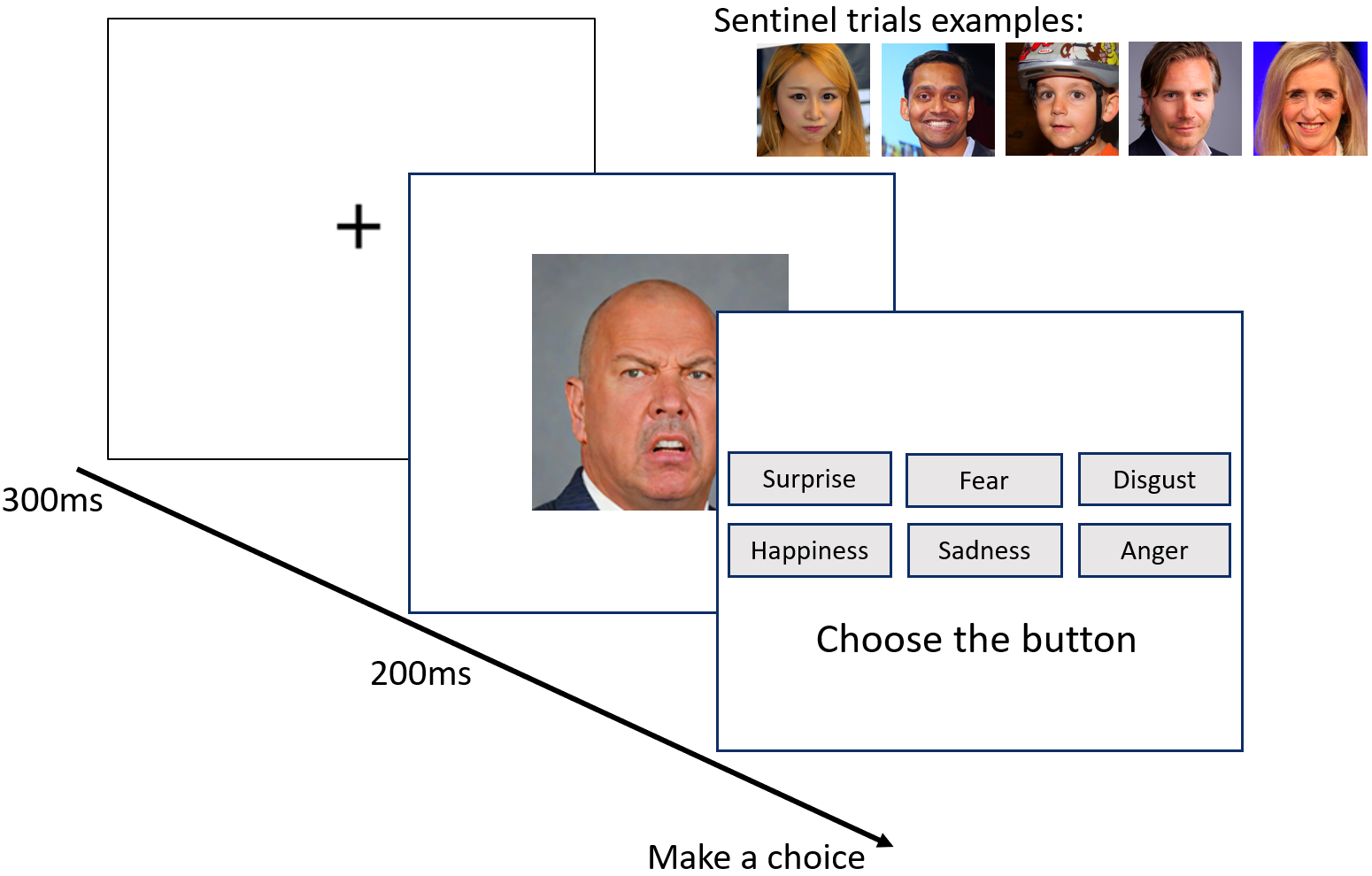}
    \caption{\textbf{Human facial expression recognition experiment procedure.} In each round of the experiment, the participant will first see a cross at the center of the screen for 300 ms. Following this, a facial image containing a specific expression will be presented for 200 ms. Next, a choice page with six buttons will appear, and the participant is required to judge the expression of the face in the image just shown and select the corresponding button. After the participant clicks a button, the current trial ends, and the next trial begins. Each participant will complete a total of 400 trials, which include 390 random trials and 10 sentinel trials. 
}
    \label{figHumanExp}
\end{figure}

Inspired by the two-stage generation method proposed in DALL-E 2 \cite{ramesh2022hierarchical}, where image embeddings are first generated and then used to create the image, we also adopt a two-stage approach for image generation. In the first stage, we apply a diffusion model to add noise and remove noise from the input image embeddings, simultaneously implementing guidance in this process: uncertainty guidance. The goal of uncertainty guidance is to sample from the perceptual boundaries of the classifier. The loss function for uncertainty is as follows:

\begin{equation} loss(x,y) = -p(y|x)*q(y) \end{equation}

where $p(y)$ is the classifier's predicted distribution, and $q(y)$ is the guiding target distribution. This loss function is inspired by GANs \cite{goodfellow2014generativeadversarialnetworks} and aims to maximize the probability of the target distribution (e.g., `fear' and `happiness') while minimizing the probability of non-target distributions, thereby generating controllable high-uncertainty images.

In previous studies, researchers attempting to use synthetic images to investigate human and model cognition often encountered the issue of unnatural or unrealistic generated images(\cite{gaziv2024strong},\cite{veerabadran2023subtle},\cite{golan2020controversial},\cite{feather2023model}). This made it difficult for participants to recognize the images, severely impacting the effectiveness of human experiments. Recent research has shown that using diffusion models as regularizers can introduce natural image priors during the generation process(\cite{jeanneret2023adversarial},\cite{wei2024cocog2controllablegenerationvisual},\cite{chen2023advdiffuser},\cite{vath2023diffusion},\cite{atakan2023dreamr}), making the images significantly more natural and realistic, which in turn helps evoke the intrinsic variability in human perception. Based on this, we employed a two-stage diffusion process to generate the final images. This approach ensures that the generated images better align with the real distribution of natural images, effectively enhancing their impact on human perception. The function for sampling process with diffusion model in first stage is as follows:

\begin{equation}
    x_{t-1} = DDPM^-(x_t)-\gamma\nabla_{x_t}loss(x_t,y)
\end{equation}

where $DDPM^-(x_t)$ represents the reverse diffusion step, $loss(x_t,y)$ is the uncertainty loss, and $\gamma$ is the hyperparameter of the guidance strength. In our experiment, we chose U-ViT\cite{bao2023worthwordsvitbackbone} as the diffusion model in the first stage generation, $\gamma=0.5$, and stable diffusion XL in the second stage generation.

\subsection{Filtering Generated Images}

Since we applied guidance only during the first-stage diffusion process, the images generated in the second stage exhibit considerable randomness. To ensure that the generated images align with the expected distribution, we filtered the images based on specific criteria. The filtering criterion is:

\begin{equation} 
p_{emotion1}>k_{emotion1}\;\&\;p_{emotion2}>k_{emotion2} 
\end{equation}

where $p_{emotion1}$and $p_{emotion2}$ are the activation values of the two emotions (emotion1, emotion2) predicted by the ANN for the generated image, and $k_{emotion1}$ and $k_{emotion2}$ are the 75th percentiles of the activation values for the corresponding emotions (emotion1, emotion2) across all images in the RAF-DB dataset. Through this filtering process, we ensure that the generated images effectively induce perceptual variability in the ANN. Activation values distribution of RAF-DB dataset can be found in Figure~\ref{figA2}.

\section{Collecting Human Perceptual Variability}

We used the filtered synthetic images as experimental materials, with the aim of collecting human participants' choices regarding these images. The experiment was approved by the local university's ethics committee before its commencement. The experiment was implemented using jsPsych and conducted online through the NAODAO platform, with a total of 100 participants. Prior to participating in the experiment, each participant read an informed consent form detailing the potential risks. Participants were free to withdraw from the experiment at any time, and no personal information was collected. During data processing, we retained feedback data from 66 participants with sentinel trial accuracy greater than 70\%, resulting in a total of 1,678 images and 22,450 trials. We used the retained participant data to construct a facial expression dataset with high perceptual variability, named varEmotion.

\textbf{Evaluation metrics.}
To effectively evaluate the efficacy of guidance in generative methods, we propose three types of guidance outcomes, as illustrated in Figure~\ref{fig3}(a): \textit{success}, \textit{bias}, and \textit{failure}. For the guiding targets $emotion1$ and $emotion2$, we define $p_1$ and $p_2$ to represent the probabilities of $emotion1$ and $emotion2$, respectively. If the result is $min(p_1, p_2) > 0.25$ and $p_1 + p_2 > 0.6$, it indicates that the generated stimulus leads individuals to select the guiding targets evenly, and we classify the guidance as \textit{success}. If the result is $min(p_1, p_2) < 0.25$ and $p_1 + p_2 > 0.6$, it suggests that all subjects tend to choose a specific target, and we classify the guidance as \textit{bias}. If $p_1 + p_2 < 0.6$, it indicates that the stimulus does not effectively influence the subjects' choices, and we classify the guidance as \textit{failure}.

\subsection{Quantitative Analysis of varEmotion}
\begin{figure}[htbp] 
    \centering
    \includegraphics[width=0.5\textwidth]{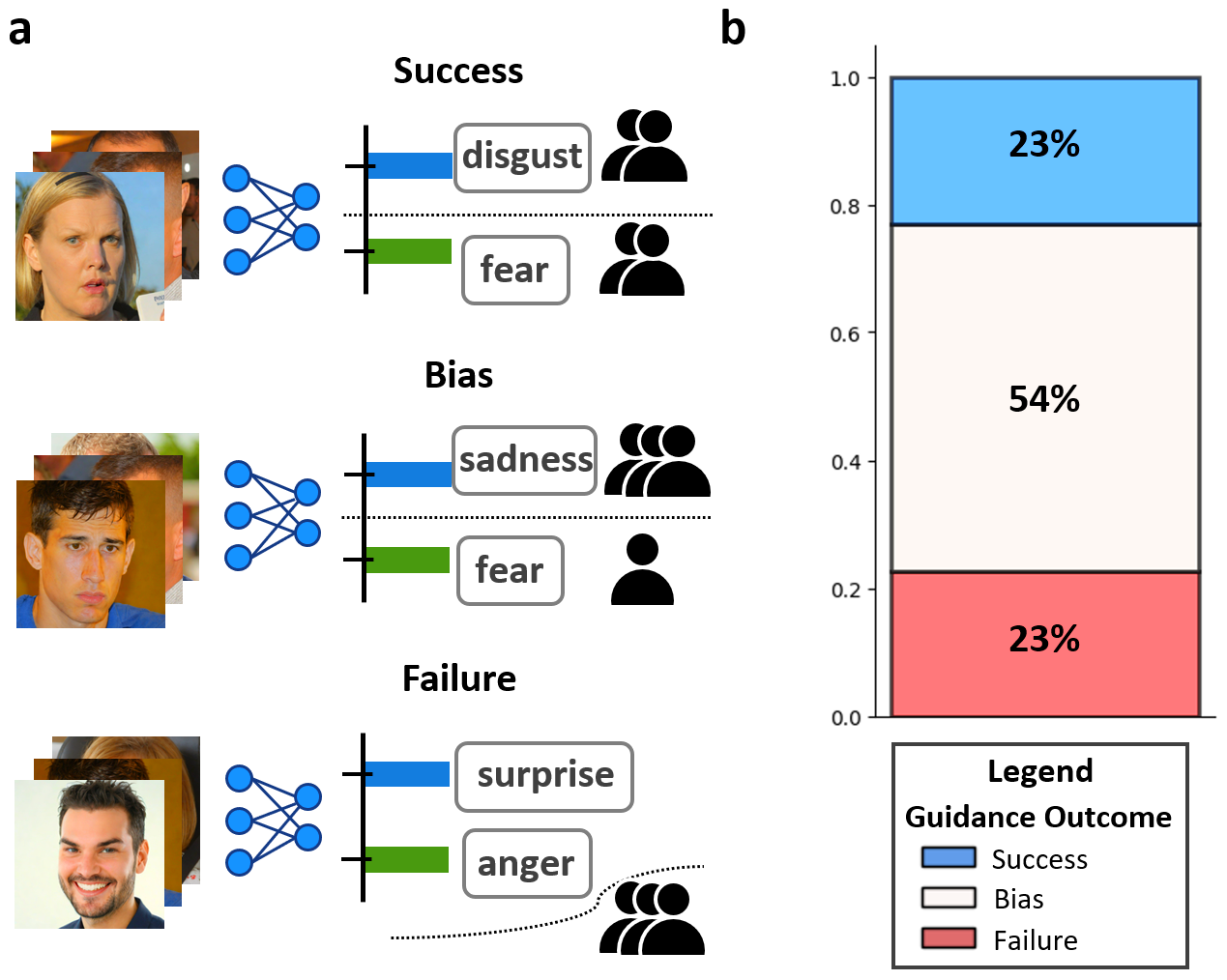}
    \caption{\textbf{Quantitative Analysis of varEmotion.} (a) Examples of three guidance outcome:\textit{success, bias, failure}. (b) Guidance outcome across the varEmotion dataset. The sum of \textbf{success} and \textbf{bias} rates approaches 80\% .
    }
    \label{fig3}
\end{figure}

\textbf{ANN variability can arouse human variability.} To examine whether the images generated by the ANN perception boundary sampling effectively evoke human subjects' perception variability, we calculated the entropy of the probability distribution of subjects' choices for emotion images and plotted the corresponding entropy distribution in Figure~\ref{figA1}. It is evident that the entropy for the vast majority of images is greater than 0, indicating that these images effectively triggered varying responses among different subjects. Furthermore, as shown in Figure \ref{fig3}(b), nearly 80\% of all generated images fall under the categories of \textit{success} or \textit{bias}. This indicates that, in the majority of cases, human choices aligned with either both or one of the guidance targets. This demonstrates that the generation method effectively guided human facial expression and emotion recognition behavior.

\begin{figure*}[htbp] 
    \centering
    \includegraphics[width=1\textwidth]{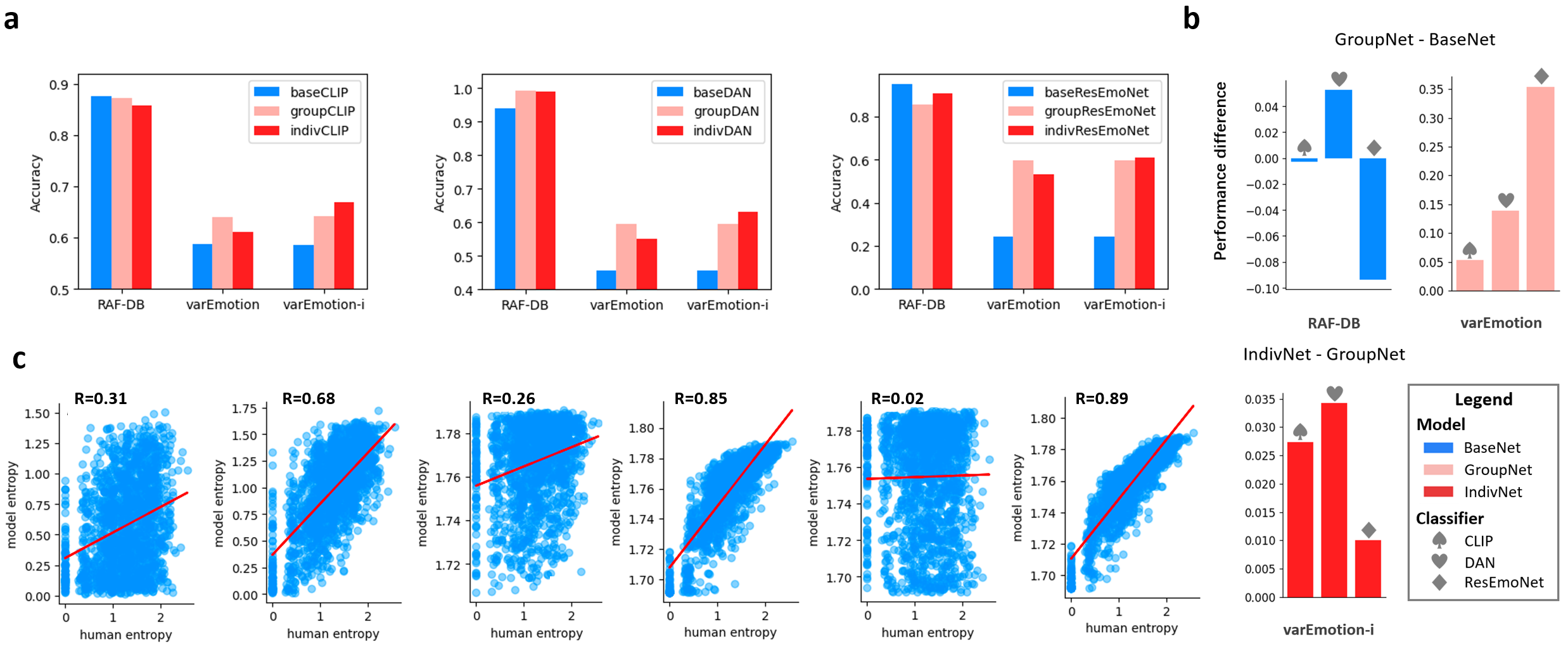}
    \caption{\textbf{Human alignment results.} (a) Accuracy of BaseNet, GroupNet and IndivNet on RAF-DB, varEmotion and varEmotion-i. On varEmotion, GroupCLIP and IndivCLIP improve 5\% over baseNet, GroupDAN and IndivDAN improve 14\%, GroupResEmoNet and IndivResEmoNet improve 35\%. On varEmotion-i, IndivCLIP outperform 2.5\% over GroupCLIP, IndivDAN outperform 3.5\% over GroupDAN, and IndivResEmoNet outperform 1\% over GroupResEmoNet. After fine-tuning different models at the individual and group levels, performance differences were observed, indicating that the model architecture is related to the model's ability to fit the boundaries of human perception. (b) Finetuning result for 3 classifiers. On varEmotion, all classifiers improved, with ResEmoNet showing the largest gains and CLIP the smallest. Individual fine-tuning further improved all classifiers with the same trend. All classifiers, after fine-tuning, were able to better predict human behavior, suggesting that fine-tuning with human data effectively enables the model to align with human perception. (c) For DAN, Spearman rank correlation between model and human entropy increased from $\rho = 0.26$ to $\rho = 0.85$ after group fine-tuning. After fine-tuning DAN at the group level, the entropy distribution of the predicted images became closer to that of human predictions, indicating that the model has captured the uncertainty in human perception.
    }
    \label{fig4}
\end{figure*}

\section{Predicting Human Perceptual Variability}
\subsection{Model Finetuning For Human Alignment}
To fine-tune CLIP for the emotion classification task, we added an MLP head at the end of CLIP and aligned it with human perception through fine-tuning the MLP head.To align models with both group-level and individual-level performance, we adopted a mixed training approach with an 4:1 split for training and validation. For individual-level datasets (varEmotion-i), the validation set was designed to avoid overlap with the group validation set. For group-level training, we combined the varEmotion and RAF-DB datasets in a 2:1 ratio, ensuring performance on RAF-DB while fine-tuning for perceptual variability. For individual-level training, we mixed varEmotion-i, varEmotion datasets in a 2:1 ratio, ensuring the models performed effectively on individual-specific and group datasets.

For group-level fine-tuning, the original classifier models were trained on mixed RAF-DB, varEmotion datasets. The pictures were normalized with `ToTensor' transformation. For Individual-level fine-tuning, the initial model is the group model. Training and testing sets were loaded with a batch size of 128, and the models were implemented with 3 different configurations to map input images to 6 output classes. Training was performed on NVIDIA GPU using Adam optimizer ($lr=1\times{10^{-4}}$) for 15 epochs, and CrossEntropyLoss function was used to compute the classification loss.


\subsection{Alignment Analysis}
\textbf{Fine-tuning improves both group-level and individual-level prediction performance.} As illustrated in Figure \ref{fig4}a, on the RAF-DB dataset, the performance of BaseNet, GroupNet, and IndivNet is comparable, suggesting that fine-tuning at both the group and individual levels did not result in a significant decrease in prediction accuracy for this dataset. In contrast, on the varEmotion dataset, both GroupNet and IndivNet show improved prediction accuracy compared to BaseNet, indicating that fine-tuning at both levels effectively enhanced the model’s ability to align with human perceptual boundaries. Moreover, on the individual dataset varEmotion-i, IndivNet, which is fine-tuned using individual data, demonstrates an average prediction accuracy improvement of 3\% over the group model GroupNet. This highlights the model’s capacity to effectively capture individual perceptual differences and better align with individual perceptual boundaries. We anticipate that in future research, IndivNet, which models individual perception effectively, will play a pivotal role in advancing our understanding of individual perception and behavior regulation.

\textbf{Different classifiers exhibit inconsistent performance.} Figure \ref{fig4}(b) compares the changes in predictive performance of various models on the RAF-DB, varEmotion, and varEmotion-i datasets after fine-tuning at both the group and individual levels. On the varEmotion dataset, the prediction accuracy of all models shows significant improvement after fine-tuning, though the degree of improvement varies across models. ResEmoNet achieves the highest accuracy gain, while CLIP shows the smallest. On the varEmotion-i dataset, accuracy continues to improve post-fine-tuning, with DAN exhibiting the greatest improvement and ResEmoNet the least. These discrepancies in accuracy changes suggest that differences may exist in the perceptual boundaries at the group, individual, and model levels, which contribute to variations in model fitting performance across different architectures.

\textbf{Human variability can be predicted by models.} To assess the alignment between model and human perceptual variability, we analyzed the correlation between model and human entropy, as shown in Figure~\ref{fig4}c. Taking DAN as an example, group fine-tuning increases the Spearman rank correlation between model and human entropy from $\rho = 0.26$ to $\rho = 0.85$. This significant improvement indicates that fine-tuning allows the model to better capture human uncertainty, aligning model predictions more closely with human perceptual behavior.

\section{Conclusion}

This study demonstrates that ANN decision boundaries serve as meaningful indicators of inter-individual perceptual variability in facial expression recognition. By leveraging a perceptual boundary sampling approach, we systematically generate stimuli that challenge both ANN classifiers and human observers, revealing a strong correspondence between machine and human perceptual uncertainty. The varEmotion dataset, constructed from large-scale behavioral experiments, provides empirical evidence that ambiguous ANN samples also evoke divergent interpretations among individuals, reinforcing the hypothesis that ANN-confusing stimuli capture key dimensions of human perceptual variability.
Beyond dataset generation, our findings underscore the feasibility of aligning ANN representations with individual-level human perceptual patterns. Through fine-tuning on behavioral data, we successfully adapt ANN models to account for subject-specific differences, highlighting the potential for more personalized affective computing systems. This research paves the way for future studies on human-machine alignment in emotion perception, suggesting that ANN decision boundaries can serve as a valuable tool for studying perceptual variability and enhancing adaptive AI-driven emotion recognition.

\section*{Acknowledgment}
This work was supported by the National Natural Science Foundation of China (62472206), Shenzhen Science and Technology Innovation Committee (2022410129, KJZD20230923115221044, KCXFZ20201221173400001), GuangDong Basic and Applied Basic Research Foundation (2025A1515011645 to ZC.L.), Shenzhen Doctoral Startup Project (RCBS20231211090748082 to XK.S.), Guangdong Provincial Key Laboratory of Advanced Biomaterials (2022B1212010003), and the Center for Computational Science and Engineering at Southern University of Science and Technology.

\appendix
In the appendix, we present supplementary figures that complement the experimental results and data analysis in the main text. Figure~\ref{figA3} illustrates a schematic diagram of perceptual boundary sampling, demonstrating how uncertainty guidance is employed to sample at the ANN perceptual boundary. Figure~\ref{figA1} provides a detailed analysis of human behavioral data, including the entropy distribution of human judgments on images, the distribution of human reaction times, and the correlation between entropy and reaction time. Figure~\ref{figA2} presents an analysis of images from the RAF-DB dataset, showing the distribution of activation values across six emotional dimensions for all images in the dataset.
\begin{figure}[!htbp]
    \centering
    \includegraphics[width=1\linewidth]{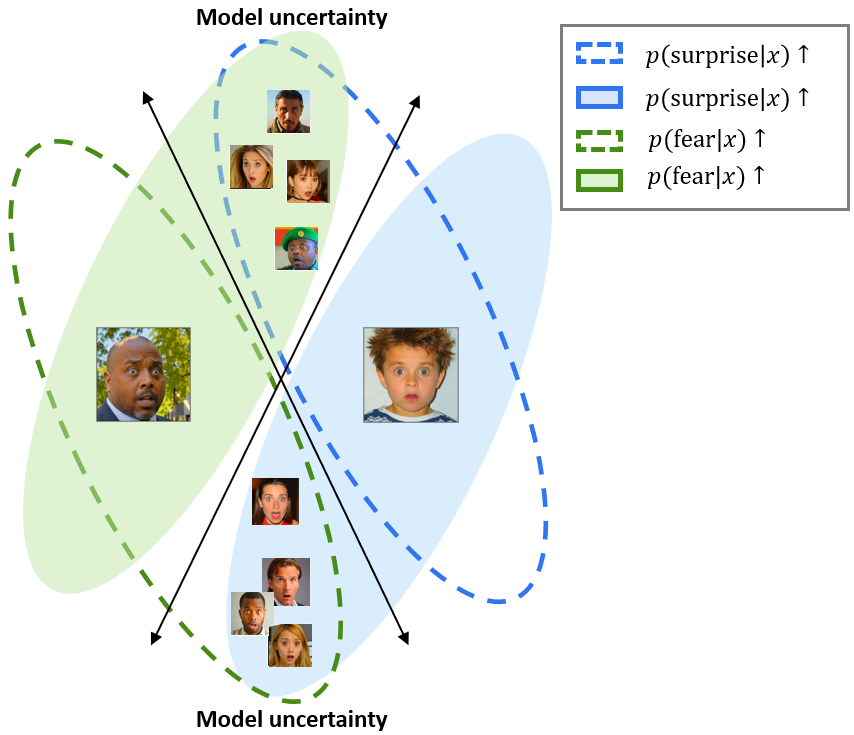}
\end{figure}
\noindent
\captionof{figure}{\textbf{Sampling on perceptual boundaries.} The perceptual space of ANN can be divided into four regions based on two classification axes. Taking the emotion pairs (fear, surprise) as an example. our objective is to generate images that induce uncertainty of ANN, as illustrated in the figure. The images in the upper and lower regions can lead the ANN to predict high probabilities for both fear and surprise. In the left region, the ANN predicts a high probability for fear but a low probability for surprise. Conversely, in the right region, the ANN predicts a high probability for surprise but a low probability for fear.}
\label{figA3}

\begin{figure}[!htbp] 
    \centering
    \includegraphics[width=0.5\textwidth]{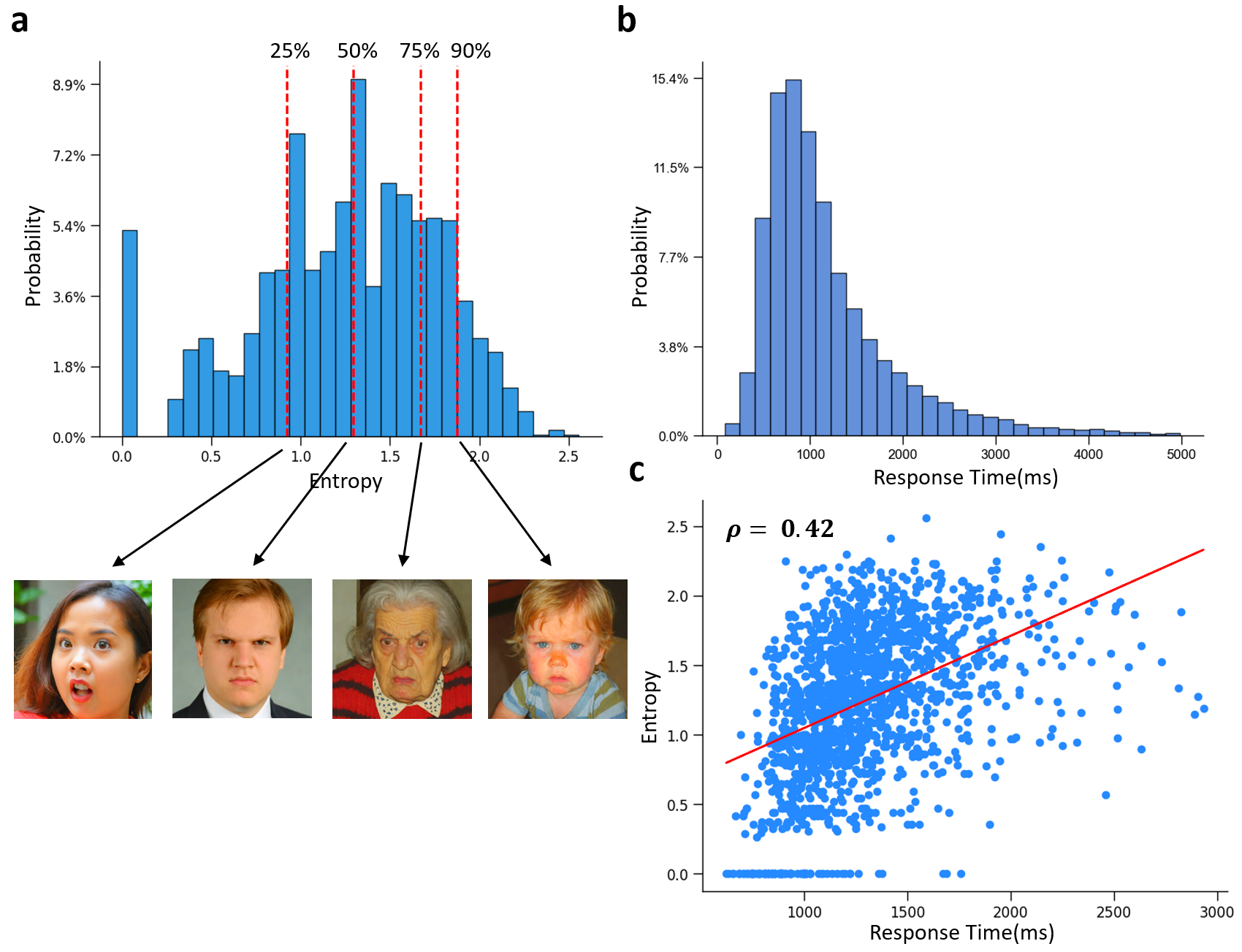}
    \caption{\textbf{Behavioral results of the Digit recognition task.} (a) The entropy distribution of human judgments on images primarily concentrates between 0.5 and 2.0, approximately following a normal distribution. (b) The time humans take to judge the images is concentrated between 500 and 1500 ms, and it exhibits the characteristics of a heavy-tailed distribution. (c) Entropy and response time exhibit a positive correlation, with a Spearman rank correlation coefficient of 0.42.}
    \label{figA1}
\end{figure}


\begin{figure}[!htbp] 
    \centering
    \includegraphics[width=1\linewidth]{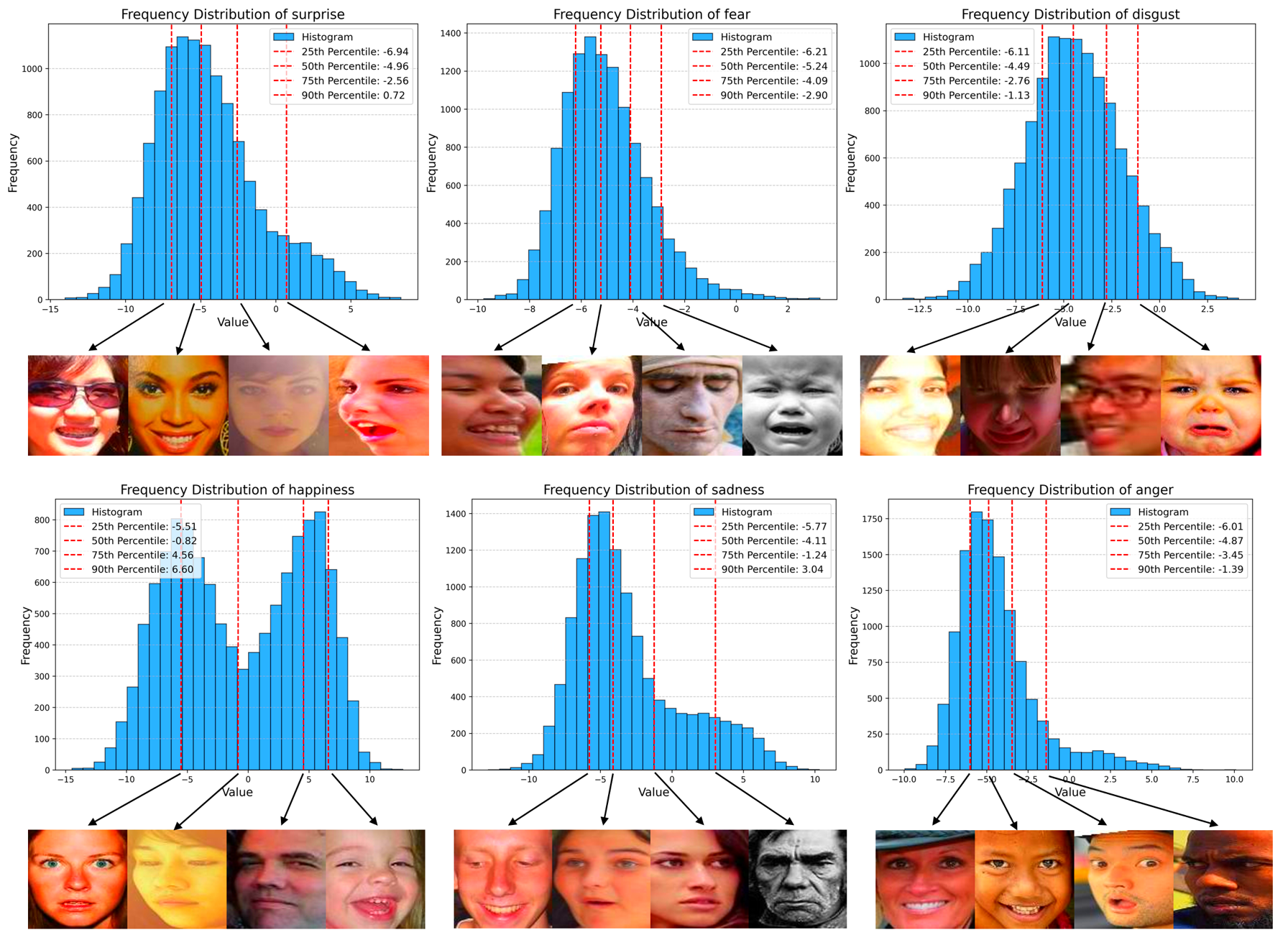}
    \caption{\textbf{Distribution of activation values varies across different emotions.} The distribution of emotion activation values in images from the RAF-DB dataset across various emotional dimensions. It can be observed that there are certain differences in the distribution of activation values for different emotions. The activation value distributions for the majority of emotions resemble a normal distribution, while the distribution for `happiness' is distinctly bimodal.}
    \label{figA2}
\end{figure}




\bibliographystyle{IEEEtran}
\bibliography{ref}

\end{document}